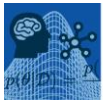

*Article*

# Machine and Deep Learning Applications to Mouse Dynamics for Continuous User Authentication

Nyle Siddiqui, Rushit Dave *, Mounika Vanamala and Naeem Seliya

Department of Computer Science, University of Wisconsin – Eau Claire, Eau Claire, WI 54701, USA; siddiqun8701@uwec.edu; seliyana@uwec.edu; vanamalm@uwec.edu
* Correspondence: daver@uwec.edu

**Abstract:** Static authentication methods, like passwords, grow increasingly weak with advancements in technology and attack strategies. Continuous authentication has been proposed as a solution, in which users who have gained access to an account are still monitored in order to continuously verify that the user is not an imposter who had access to the user credentials. Mouse dynamics is the behavior of a user's mouse movements and is a biometric that has shown great promise for continuous authentication schemes. This article builds upon our previous published work by evaluating our dataset of 40 users using three machine learning and deep learning algorithms. Two evaluation scenarios are considered: binary classifiers are used for user authentication, with the top performer being a 1-dimensional convolutional neural network (1D-CNN) with a peak average test accuracy of 85.73% across the top-10 users. Multi-class classification is also examined using an artificial neural network (ANN) which reaches an astounding peak accuracy of 92.48%; the highest accuracy we have seen for any classifier on this dataset.

## 1. Introduction

The optimal user authentication method must be flexible, computationally efficient, operational in near real-time, and most importantly, accurate. Several methods of authentication have been previously proposed with biometrics appearing to be the frontrunner for future user authentication schemes. Biometric-based authentication is segregated in two groups: physical biometrics and behavioral biometrics [1]. Physical biometrics relies on the uniqueness of certain physical attributes among humans for authentication by exploiting fingerprints [2], irises [3], and voice scanners [4]. Similarly, behavioral biometrics operates under the assumption that the general behavior of humans for certain tasks are distinct enough to be used for user authentication, with examples including touch dynamics [5], keystroke dynamics [6, 7], and the focus of this paper: mouse dynamics. Behavioral biometrics for authentication, when compared to physical biometrics, has been of particular interest due to its more general, large-scale applicability, reduced intrusiveness, and lack of external sensors. Moreover, mouse dynamics has proven to be a continuous, lightweight, and non-intrusive method for dynamic user authentication [8-11].

Further, the large amounts of data available in the domain space of mouse dynamics allows for the leverage of machine and deep learning algorithms. Fields which possess sufficiently large amounts of data have observed tremendous improvements in results when machine/deep learning methods are integrated. Machine learning exploits implicit patterns in large data too difficult for humans to detect and has shown to be robust enough to be applied to a variety of problem spaces, however it requires the manual extraction of features from data. This makes it difficult to find the optimal combination of hyperparameters, extracted features, and data pre-processing methods for a given problem or dataset. Conversely, deep learning methods utilize a hierarchy of layers to build levels of abstraction from raw data. This removes the necessity for hand-picked extracted features and allows the model to gain a more holistic understanding of the data as opposed to being limited to only the manually presented features. Despite the benefit of increased model autonomy, deep learning methods come with their own introduced difficulties, such as vanishing and exploding gradients, non-convergent loss functions, and less interpretable results – often referred to as the "black box" nature of deep learning.

Due to the relative infancy of mouse dynamics and the sparsity of publicly available data, the integration, impact, and shortcomings of these algorithms in the field have not been fully realized. Specifically, mouse dynamics has generally been approached as a method for user authentication in low-intensity settings. While promising results have been published in these settings, the full potential of mouse dynamics has not been fully utilized until it has been thoroughly tested in many different evaluation scenarios. In tandem with the use of machine and deep learning, mouse dynamics could potentially serve as an inexpensive, non-intrusive behavioral biometric for user authentication. Thus, the scope of this paper revolves around evaluating three machine learning and three deep learning algorithms on an improved novel dataset introduced in our previous work [12]. This dataset contains the mouse dynamics data of 40 users as they engaged in a high-intensity task on a desktop computer for 20 minutes. Current publicly available mouse dynamics datasets, such as the Balabit [13] and TWOS [14] datasets, were collected with small sample sizes and with dull, administrative tasks given to users to engage with during data collection. Evaluating these 6 algorithms on data collected in a more high-intensity, volatile environment allows for a more scrutinizing examination of the integration of machine and deep learning to mouse dynamics, as well as may provide an insight into the required future research to further improve the problem space. We found our 1D-CNN and ANN to be the best performing algorithms in the binary and multi-class classification scenarios, respectively, with more details found in the Results section. Thus, the novel contributions of this paper are as follows:

- Improve upon our previous work by introducing and analyzing a larger dataset (can be found in the Data Availability section)
- In addition to random forests, two new machine learning models and three new deep learning models are evaluated on the dataset
- Introduce an artificial neural network for multi-class mouse dynamics user classification

## 2. Background

This paper builds on the results of [12], where we hypothesized that collecting data in an environment where users have more variability and freedom to choose their actions would naturally encode more distinct features in each user's data. We proceeded by collecting data from 10 users while they played their own 20-minute game of *Minecraft* on a designated lab computer, then trained binary random forest classifiers for user authentication. Upon any mouse event, (i.e. mouse movement or mouse click) a custom program running in the background collected the UNIX timestamp of the action, X and Y coordinates of the pointer, and corresponding Subject ID. Raw user data, visualized in Figure 1, is not sufficiently granular for machine learning algorithms to reliably authenticate users, hence the need for the aforementioned data processing and feature extraction stages. Ten singular mouse events were combined into a singular block called a "mouse action" which is what is inputted into the random forest for classification. Note that variable mouse action block lengths have also been proposed, with mouse action block lengths usually a great order of magnitude larger than 10. However, using a smaller mouse action block length and observing similar results indicates that our dataset encodes more information in a smaller amount of data, which is what we were able to conclude in [12]. Additional features such as velocity, acceleration, jerk, angular velocity, angle and length of mouse trajectory, along with their summary statistics, were extracted from each mouse action and placed into a vector along with the target class associated with the action. Ten binary random forests for user authentication were correspondingly trained to each user, with inputted data consisting of an equal number of positive and negative class instances to balance each classifier. A positive instance was defined as a genuine action that should warrant continued user access, while negative instances were imposter mouse actions that should result in additional monitoring/blocking of the user's access.

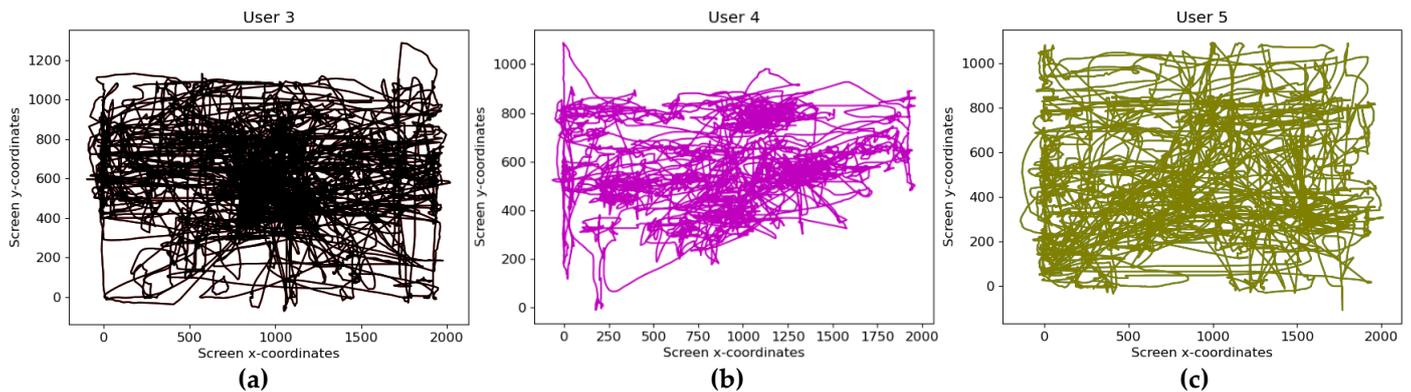

**Figure 1.** (**a**) Visualization of raw mouse from User 3's session; (**b**) Visualization of raw mouse from User 4's session; (**c**) Visualization of raw mouse from User 5's session.

Many data processing and model building decisions in [12] were inspired by the original paper proposing mouse dynamics as a reliable behavioral biometric [15]. Authors of [15] introduced mouse dynamics as a novel human-computer interaction behavior that could serve as an alternative security medium. As opposed to conventional methods, where security relies on something "one knows" (PINS, passwords) or "one has" (keys, identification cards), mouse dynamics offers security through what "one is." In other words, irreproducible behavioral biometrics such as mouse dynamics are difficult to mimic and thus are a great candidate for user authentication schemes. Similarly to [16], the authors of [15] do note that due to the variability in human behavior over time, behavioral biometrics as a whole must be designed to be robust and dynamic to changes in user behavior. [15] goes on to establish baseline data preprocessing and evaluation methods, such as what additional features to extract from raw data, the use of false positive, false negative, and equal error rates as evaluation metrics, the combination of mouse actions into variable length mouse action blocks, and even offers results using statistical learning methods to compare for future research. Despite the recency of mouse dynamics, great improvements have been made to the field since [15]. Most noticeably, the consistent superior performance machine learning has exhibited over more traditional statistical learning methods is also quite apparent in mouse dynamics [17]. Moreover, [17] proposed additional extracted features for mouse dynamics-based authentication and reported their gain ratios. The search for new and more descriptive extracted features compared to [15] and the magnitude of their impact on accuracy is imperative to the improvement of the field and has been heavily researched [18, 19], however can also be seen as a limitation for the application of machine learning; even more so when robust evaluation techniques and metrics are encouraged [20]. This is observed to an even greater degree when deep learning has been applied and performed at a higher level without the need for feature extraction [21, 22] - presumably due to deep learning's ability to perform at superhuman levels with sufficiently large data.

Deep learning's performance on mouse dynamics can be dependent on the approach to the problem. For example, due to the spatiotemporal attributes of the data, mouse dynamics can be framed as a time-series problem yet integrate deep learning through different modalities. There are multiple ways to leverage CNNs for mouse dynamics [23], such as using the X and Y coordinates from the raw data to create user mouse maps, as observed in [12], and evaluate their spatial features using a two-dimensional CNN [24]. By focusing on and extracting the spatial features of mouse dynamics data, the image processing capabilities of CNNs can be similarly applied to user authentication. However, the strengths of convolutions in CNNs extend beyond spatial features, as the temporal approach of modeling data as a time-series signal and using a 1D-CNN to convolve along the singular time dimension of the signal has been proposed as well [25]. Further, [25] uses X and Y velocities to create signals from their mouse events rather than absolute coordinates to produce a translation invariant model. Consequently, recurrent neural networks – especially their long short-term memory variants (LSTM-RNN) - have been proposed for use due to their known strong performances with time-related data [26, 27]. The contrasting methods in which these models exploit the spatiotemporal features of mouse dynamics data could have an impact on their performance, hence their inclusion in our experiment.

It is important to note that two-class binary classifiers have been shown to outperform one-class classifiers on behavioral biometrics as a whole [28]. To ensure a more holistic evaluation of this dataset, we also implement a singular model to distinguish between all users in a multiclass classification setting. Binary user authentication is only concerned with identifying whether a mouse movement originated from a genuine user or an imposter, however this only evaluates a model's ability to differentiate between inter-class instances (i.e., imposter vs. genuine user). The ability to further distinguish one genuine user from another on a trusted network (intra-class instances) possesses great potential value and should not be ignored. As such, this paper introduces an artificial neural network to operate under such constraints in mouse dynamics, yielding the question of whether more complex deep learning algorithms may be able to follow suit with further research.

**3. Methodologies**

The data preprocessing methods differ by model. Machine learning models require manual feature extraction, whereas the deep learning models were simply given raw mouse dynamics data for implicit feature extraction and analysis. Due to the architectural differences between models, the various spatial and temporal aspects of the data are leveraged differently and at an unequal frequency dependent on the model; for example, the aforementioned difference between an LSTM-RNN and 1D-CNN. Furthermore, across both deep and machine learning applications to mouse dynamics, binary classifiers have been the most commonly used approach. This approach revolves around creating a separate classifier for each user. Each user's dataset consists of an equal number of positive and negative class instances. This is usually achieved by sampling all available instances from the target user (positive class) and appending an equal number of imposter instances (negative class). The imposters instances are simply random samples from the rest of the remaining users who are not the target user. In a dataset with $k$ users, there should be $n$ positive and negative instances in a user's dataset where $\frac{n}{k-1}$ imposter samples are taken from each non-target user. This creates a balanced dataset that mitigates frequency bias in the classifiers. Therefore, instead of tasked with distinguishing each user from one another, a classifier is only tasked with identifying a mouse action as a genuine action or an imposter action. Our feature extraction process extracted 33 additional features from each user's raw data, such as velocity, acceleration, jerk, curvature, and trajectory values. More details regarding our preprocessing method for the machine learning algorithms can be found in our previous paper [12].

While quite frequent in mouse dynamics, binary classification prevents more nuanced inferences on the data, such as which users share the most similar behaviors. Therefore, we also include a multi-class classification approach to mouse dynamics, which comes with its differences from the usual methods previously proposed. This section serves to describe our methodologies as well as the few but important dissimilarities between models and their preprocessing methods. All models were built and evaluated in Python using the Keras/TensorFlow and the scikit-learn libraries.

*3.1. Deep Learning Models*

All deep learning models and their respective layers featured in this paper are available in the TensorFlow and Keras libraries.

3.1.1 LSTM-RNN

RNNs in general have seen their fair share of exposure in the mouse dynamics literature [29, 30] and with the advent of LSTM-RNNs, continue to produce excellent results across many research domains. Mouse dynamics, generally, can be represented as a sequence of distinct actions over a variable period of time, thus making LSTM-RNNs a strong candidate as a model choice time-series related data. As a part of the family of neural networks, LSTM-RNNs excel at taking some sequence of inputs, propagating this data through the network to build higher levels of abstraction, and producing a meaningful output. Specifically, LSTM-RNNs have the added benefit of internal memory, or the capability to identify long-term dependencies. This benefit has been more readily realized in areas such as natural language processing due to the intertwined and dependent nature of linguistics, however these benefits do apply to general time-series data as well. This requires the input to be in the form of a 3-dimensional array, with the dimensions representing batch size, number of features, and number of time steps. Therefore, the preprocessing sequence for our LSTM-RNN consisted of creating one block of mouse actions using 10 mouse events, similar to [12]. Due to the static length of 10 imposed on each mouse action, as opposed to variable length proposed in other literature, a time step of 3 was used for the LSTM-RNN. We hypothesized that using a timestep of 3 would encode additional information into each data point passed to the LSTM-RNN while still leaving a sufficient amount of data to train a model of our respective size.

The architecture of the model consists of an input layer, three hidden LSTM layers with 128 neurons, followed by a dropout layer and a batch normalization layer, as is standard in most LSTM-RNN architectures. The hidden layers are followed by a single fully connected layer of size 64 with a ReLU (Rectified Linear Unit) activation, followed by a final fully connected layer of size 1 with a softmax activation for classification output. The Adam optimizer is used with $\alpha$ = 0.001 with a decay rate of $1.0 * 10^{-6}$. Binary cross-entropy loss is used due to the binary nature of the data (whether the data is from the target user or not).

3.1.2 1D-CNN

Mouse dynamics data can be represented through many modalities. As opposed to examining long-term behavioral tendencies using just time, 1D-CNNs additionally leverage both the spatial and temporal nature of the data. CNNs in general have observed state-of-the-art performance in all facets of computer science, most notably computer vision. These models are built on the foundation of processing data with a grid-like topology using convolutional arithmetic rather than using general matrix multiplication for computation within layers. With regards to CNNs specifically, liberties have been taken with the representation of data for mouse dynamics, as an optimal representation has not yet been found due to the novelty of the field. Some suggestions include using speed rather than absolute coordinates to ensure a translationally invariant model [25] or graphing mouse movements on a 2-dimensional plane and using a traditional 2D-CNN [24]. Many of these methods are compatible with our data, as Figure 2 outlines. As such, we incorporate using speed values for mouse action sequences rather than raw coordinate values to create a more robust and flexible model. Our mouse dynamics data can be represented as a one-dimensional time series signal and thus is suitable for analysis by a 1D-CNN. Similarly to our LSTM-RNN, data is inputted as a 3-dimensional matrix. However, instead of using a timestep of 3, we simply pass a single mouse action as input to the model. Since an action already consists of 10 events, the model will convolve over each event examining the X and Y speed values at each UNIX timestamp and thus not requiring any batched timesteps.

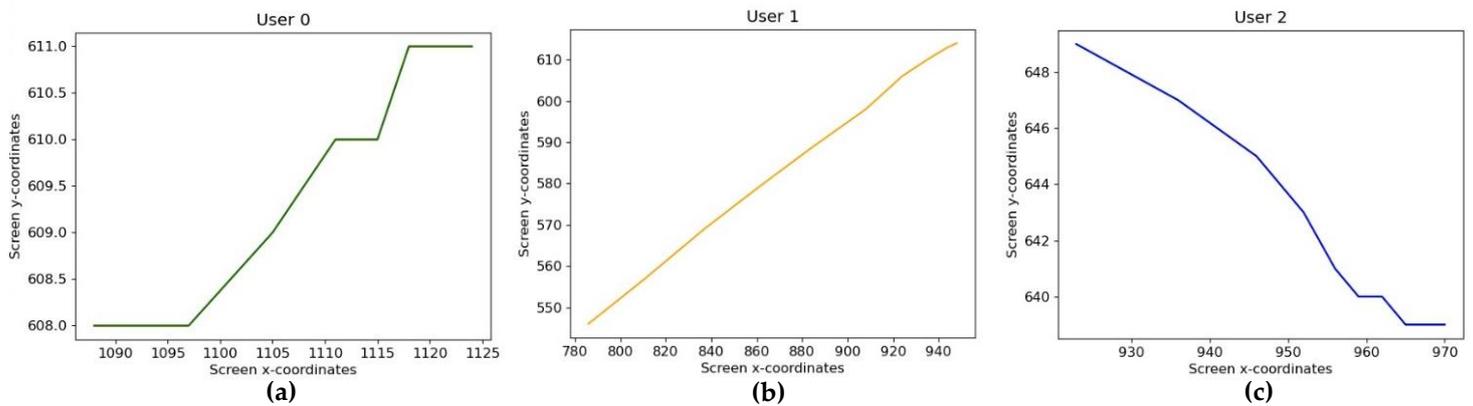

**Figure 2.** (**a**) Visual representation of one of user 0's mouse action with block length 10; (**b**) Visual representation of one of user 1's mouse action with block length 10; (**c**) Visual representation of one of user 2's mouse action with block length 10.

The architecture of the model follows closely to [25], however is shallower in width. Model sizes of varying capacities were tested, however yielded negligible improvements in overall accuracy. A binary classification approach identical to the one explained in section 3.1.1 is taken. Thus, our model is comprised of two 1D convolutional layers, followed by a global max pooling layer. A flatten layer follows to mutate the output of the max pooling layer into a vector, leading to two fully connected layers: one layer with a size of 60 and a ReLU activation, and another layer with a size of 1 and a softmax activation for binary classification. The same loss functions and hyperparameters are used as discussed in section 3.1.1.

3.1.3 ANN

Almost all deep learning applications to mouse dynamics have operated under the constraint of binary classification. While this method of authentication has its benefits, a multi-class approach allows for more granular results. Continuous user authentication using mouse dynamics has a broader applicability outside of just intrusion detection and consequently, certain situations may arise when multi-class classification schemes are more appropriate. For example, the ability to monitor and distinguish between known users on a large network is only possible when models have been proven to operate in a multi-class setting. Thus, it is integral to explore the other settings in which mouse dynamics can reliably operate as a behavioral biometric for user authentication. Problems that could advance the biometric and the field of continuous user authentication as a whole can only be unearthed if mouse dynamics is continued to be thoroughly and wholly researched. Therefore, we propose an ANN that can distinguish between all 40 users simultaneously in a multi-class classification setting.

ANNs are one of the simpler manifestations of the neural network, inspired by the modern hypothesis of how biological neural networks operate. Raw input propagates through the network as it is manipulated by a series of affine non-linear transformations in the form of multiplicative weights and additive biases. As ANNs are trained in a supervised setting, the difference between model predictions and target values can be represented as a loss function and minimized using some variation of a first-order iterative optimization algorithm, most commonly gradient descent. ANNs are also universal approximators and thus are capable of estimating infinitely complex functions. It then follows that they are a fundamental tool in the novel application of deep learning to a field as a baseline reference.

The model follows a simple architecture, comprised of two layers of size 256, two layers of size 128, two layers of size 64, and a final output layer of size 40. All layers use a ReLU activation except for the output layer, which uses a softmax activation. Data preprocessing methods differ only slightly from the 1D-CNN and LSTM-RNN processes. A singular mouse action is inputted at a time, but raw absolute coordinates are used instead of speed values since translation invariance is not applicable in this setting. Sparse categorical cross-entropy loss is used instead of binary cross-entropy due to the multi-class classification setting.

*3.2 Machine Learning Models*

All machine models featured in this paper were built using the Scikit-learn library, specifically the sklearn module. KNNs and RFs are available in the "ensemble" and "neighbors" sklearn modules, respectively.

3.2.1 SVM

Support vector machines (SVM) are a machine learning algorithm commonly used for classification and regression problems. In essence, SVMs create hyperplanes (also known as decision boundaries) that best separates distinct classes in the data. Because of the multidimensional nature of complex ML problems, it can be difficult to find such clear separation of classes. In general, SVMs operate by finding the hyperplane that maximizes the separation between classes in an $n$-dimensional space. In simpler settings, SVMs often utilize linear classification sufficiently enough to come to a classification decision, however this may not always be the case. Similarly to the theoretical foundations of neural networks, non-linear transformations tend to yield better results when working with complex, high-dimensional data - as often is the case with mouse dynamics data and most machine learning problems. Classes may be more easily separable when mapped to higher dimensions, however explicitly mapping each sample vector to a higher dimension may be computationally inefficient and cumbersome. Kernel functions are used by SVMs to bypass this problem by acting as a similarity function for raw data representations, as opposed to explicitly computing distances by mapping sample vectors into a desired feature space. This allows for SVMs to feasibly operate in higher dimensional spaces without much of an increase in computational complexity; an integral attribute for a user authentication model based on mouse dynamics. Furthermore, SVMs usually operate in a binary classification setting. While multi-class SVMs have been proposed in the form of "one vs. rest" or multiple aggregated "one vs. one" scenarios, this article utilizes SVMs as general binary classifiers to be consistent with the other machine learning models tested. While SVMs are excellent at operating in higher dimensions and being memory efficient, they are still quite sensitive to noise and are also hard to interpret. Because classification decisions are made using a hyperplane separation, there are no probabilistic aspects that offer insight into the magnitude of error the model is experiencing. In the context of mouse dynamics, this could prove to be a hindrance in the widespread adoption of SVMs for commercial use, as identifying model weaknesses and troubleshooting would be more ambiguous of a process as compared to other alternative models. Lastly, hyperparameter optimization in machine learning models is just as imperative of a process as in deep learning, but the decreased model complexity of most machine learning models lead to a fewer number of hyperparameters needing to be tweaked. The kernel function of an SVM is arguably the most important "hyperparameter", so we use a radial basis function kernel due to its general applicability to most machine learning problems and its easy interpretability of distance using an l2-norm.

3.2.2 KNN

The k-nearest neighbor algorithm (KNN) is a nonparametric statistical learning algorithm commonly used for classification and regression problems. KNNs operate under the general assumption that similar examples will exist in close proximity to one another. Assuming this, KNNs will calculate the distance of an unknown point to other previously labeled groups. The algorithm will classify this unknown point as a member of the group if there is a sufficient number of neighbors, *K*, that are in close proximity, hence the need for a distance function. Similarly to SVMs, l2-norms are usually the most common choice for KNN algorithms but are not always necessary.

In contrast to SVMs, KNNs have been shown to be highly robust against noise and perform better on larger datasets. Due to the nonparametric, instance-based learning techniques of KNNs, they also do not require any prior training on a dataset before creating classification predictions. Consequently, there is only one hyperparameter to tune: *K*, the number of neighbors that an example must be in close proximity to in order to be associated with that group. Unfortunately, optimizing this hyperparameter is quite difficult, and after using a cross-validated random search for hyperparameters, we found that setting *K* equal to 13 yielded the best general results across most users. As with most statistical learning methods, KNNs struggle to yield high accuracy rates with higher-dimensional data. They also require feature scaling, such as normalization of the data, and are quite sensitive to outliers. The contrasting strengths and weaknesses of SVMs and KNNs allows us to test these models in a pseudo-adversarial environment and determine which aspects of our data (dimensionality, size of dataset, number of outliers, etc.) have a stronger impact in final accuracy results.

3.2.3 Random Forest

Random forests (RF) are an ensemble training algorithm, consisting of an amalgamation of decision trees that use majority voting to curate a classification or regression prediction. They are quite popular due to their simplicity, flexibility, and interpretability. At their core, RFs are a massive collection of if-else statements. At each node of a decision tree in an RF, a "test" is conducted on the inputted value. Each branch at that node denotes the total possible sample space for the inputted value. As the inputted value descends down the RF, data is granularly separated and further categorized by the various decision trees. It is hypothetically possible to visually graph each and every decision tree in a RF, making their decision-making process more transparent and easier to understand than almost any other algorithm. Furthermore, differing magnitudes of each feature does not impact the ability for RFs to perform, thus rendering feature scaling as irrelevant. RFs are known to be quite computationally cheap when evaluating test data, however the cost of evaluation for RFs are a stark contrast to their cost of training. Training on a large data set may require an extremely deep RF as it will need to develop a myriad of decision trees to deal with all possible scenarios. Thus, almost all of the computational cost resides in the training of the model rather than the actual use. RFs are actually quite efficient and low cost when classifying data, it is simply the model development that is costly.

There are many trade-offs to consider in an ensemble training algorithm. The main causes of errors in most learning models are noise, bias, and variance. So, any training method that can best mitigate these errors have a high chance to improve the accuracy and strength of machine learning algorithms. Ensemble training is designed to combine multiple decisions from multiple models to minimize the errors from each model. In the previously discussed algorithms, they are structurally independent and solely rely on a decision from a single model to classify a data point. Ensemble training allows multiple weak classifiers to be grouped into a family, and the collective decision of these weak classifiers often times produce a moderate, averaged output. Ensemble training operates under the assumption that each family of weak classifiers is going to produce errors, and understandably so. When making groups of weak classifiers, it is done so such that each group's correlation between one another is minimized. Thus, one group's proneness to a certain error may not be as prevalent in the other groups. An example of this in the context of mouse dynamics has already been observed in [31] Combining the decisions from all of the classifiers then minimizes each other's errors and create a singular strong classifying method that is unlikely to overfit. With regards to mouse dynamics, RFs serve as a reliable middle ground between the stark contrasts of KNNs and SVMs. Evaluating the differences in evaluation metrics across these three models may allow us to gain more insight on which models perform better on mouse dynamics for user classification. After performing an extensive grid search for optimal hyperparameters, we found that the following settings for a Keras random forest produced the best general results: *n_estimators* = 1600, *min_samples_split* = 2, *min_samples_leaf* = 1, *max_features* = "sqrt", *max_depth* = 30, and *bootstrap* = "False". For clarity, all hyperparameters for each model (or optimizer hyperparameters for the deep learning models) are listed in Table 1.

**Table 1.** List of hyperparameters for each model

| Model | Hyperparameter | Value |
|---|---|---|
| 1D-CNN/LSTM-RNN/ANN | $\alpha$ (Learning rate) | 0.001 |
| | $\epsilon$ (Decay rate) | 0.000001 |
| KNN | K-neighbors | 13 |
| SVM | C | 1 |
| | kernel | "rbf" |
| RF | num_estimators | 1600 |
| | min_samples_split | 2 |
| | min_samples_leaf | 1 |
| | max_features | "sqrt" |
| | max_depth | 30 |
| | boostrap | "False" |

**4. Results**

To stay consistent with results in [12], we evaluated each binary classifier using four different metrics: accuracy (ACC), false positive rate (FPR), and false negative rate (FNR). In abstract evaluation scenarios, FPR and FNR can be seen as equally erroneous, however FPRs are much more detrimental in a user authentication scenario. Explicitly, failing to authenticate a user poses a smaller security risk than continuing to authenticate an imposter. For brevity's sake, we only include FPR when evaluating the deep learning models. EER denotes the equilibrium point between FPR and FNR and is used in [12, 32] as a more general metric to compare different classifiers., but we have decided to use the F1-score - the harmonic mean of the precision and recall – in this study as it better describes the model's precision and recall capabilities [33]. Furthermore, reporting individual results for each user in this article is spatially infeasible, so we resort to reporting the results and averages from the top-10 users in accuracy (only applicable for the binary classifiers). ANN results are gathered using the entire dataset without the need to separate users into binary classes and thus is more reflective of how a baseline model would perform on every user in this dataset. All results were compiled and visualized using the "metrics" module in the sklearn library. The reported results from each model sorted by accuracy can be seen below in Tables 2-7 with the best and worst values being highlighted in green and red, respectively:

**Table 2.** 1D-CNN results

| User | ACC | FPR | F1 Score |
|---|---|---|---|
| 13 | 0.8637 | 0.1563 | 0.9197 |
| 24 | 0.8634 | 0.1602 | 0.9202 |
| 2 | 0.8587 | 0.1839 | 0.8551 |

| 39 | 0.8581 | 0.1588 | 0.9181 |
| 15 | 0.8577 | 0.1607 | 0.9225 |
| 30 | 0.8575 | 0.1194 | 0.9191 |
| 19 | 0.8567 | 0.1513 | 0.9171 |
| 14 | 0.8525 | 0.1652 | 0.8934 |
| 0 | 0.8524 | 0.1488 | 0.9182 |
| 9 | 0.8521 | 0.1419 | 0.9151 |
| **Average** | 0.8573 | 0.1546 | 0.9099 |
| **Standard Deviation** | 0.004156 | 0.01668 | 0.02091 |

**Table 3.** LSTM-RNN results

| **User** | **ACC** | **FPR** | **F1 Score** |
|---|---|---|---|
| 19 | 0.8613 | 0.1553 | 0.8948 |
| 7 | 0.8589 | 0.1564 | 0.9052 |
| 24 | 0.8562 | 0.1457 | 0.8948 |
| 6 | 0.8561 | 0.1569 | 0.8920 |
| 38 | 0.8560 | 0.1515 | 0.9110 |
| 30 | 0.8554 | 0.1659 | 0.9142 |
| 15 | 0.8548 | 0.1511 | 0.9174 |
| 37 | 0.8525 | 0.1482 | 0.9084 |
| 33 | 0.8525 | 0.1737 | 0.9136 |
| 12 | 0.8526 | 0.1685 | 0.8836 |
| **Average** | 0.8556 | 0.1561 | 0.9035 |
| **Standard Deviation** | 0.002849 | 0.008327 | 0.01141 |

**Table 4.** SVM results

| **User** | **ACC** | **FPR** | **FNR** | **F1 Score** |
|---|---|---|---|---|
| 4 | 0.6518 | 0.1692 | 0.6024 | 0.6541 |
| 39 | 0.6417 | 0.3996 | 0.4408 | 0.6426 |
| 10 | 0.6292 | 0.3964 | 0.4718 | 0.6256 |
| 0 | 0.6174 | 0.4325 | 0.4718 | 0.6206 |
| 32 | 0.6066 | 0.4362 | 0.3013 | 0.5930 |
| 5 | 0.6037 | 0.2954 | 0.6511 | 0.6031 |
| 1 | 0.5890 | 0.2875 | 0.6158 | 0.5895 |
| 3 | 0.5841 | 0.4112 | 0.4238 | 0.5832 |
| 15 | 0.5741 | 0.2361 | 0.7013 | 0.5873 |
| 31 | 0.5730 | 0.2818 | 0.6191 | 0.5662 |
| Average | 0.6071 | 0.3345 | 0.5299 | 0.6065 |
| **Standard Deviation** | 0.02771 | 0.09274 | 0.1259 | 0.02819 |

**Table 5.** KNN results

| User | ACC | FPR | FNR | F1 Score |
|---|---|---|---|---|
| 8 | 0.6584 | 0.4994 | 0.4030 | 0.6670 |
| 35 | 0.6536 | 0.4260 | 0.4306 | 0.6515 |
| 12 | 0.6195 | 0.3697 | 0.3311 | 0.6192 |
| 6 | 0.6190 | 0.4754 | 0.3118 | 0.6281 |
| 34 | 0.6066 | 0.4028 | 0.4744 | 0.6026 |
| 32 | 0.6054 | 0.3650 | 0.2484 | 0.6054 |
| 30 | 0.6033 | 0.4610 | 0.4165 | 0.6064 |
| 11 | 0.6021 | 0.4741 | 0.4045 | 0.6114 |
| 1 | 0.5938 | 0.4745 | 0.3558 | 0.5936 |
| 39 | 0.5932 | 0.4327 | 0.4102 | 0.5921 |
| **Average** | 0.6155 | 0.4380 | 0.3786 | 0.6177 |
| **Standard Deviation** | 0.02309 | 0.04683 | 0.06647 | 0.02465 |

**Table 6.** RF results

| User | ACC | FPR | FNR | F1 Score |
|---|---|---|---|---|
| 8 | 0.6827 | 0.2582 | 0.3764 | 0.7016 |
| 12 | 0.6740 | 0.3076 | 0.3442 | 0.6739 |
| 35 | 0.6580 | 0.2732 | 0.4108 | 0.6564 |
| 11 | 0.6575 | 0.3692 | 0.3158 | 0.6673 |
| 6 | 0.6548 | 0.3171 | 0.3734 | 0.6745 |
| 34 | 0.6507 | 0.2820 | 0.4165 | 0.6501 |
| 32 | 0.6408 | 0.3923 | 0.3259 | 0.6401 |
| 30 | 0.6355 | 0.3761 | 0.3569 | 0.6575 |
| 1 | 0.6305 | 0.2454 | 0.4937 | 0.6247 |
| 18 | 0.6212 | 0.4022 | 0.3553 | 0.6210 |
| **Average** | 0.6506 | 0.3223 | 0.3768 | 0.6567 |
| **Standard Deviation** | 0.01912 | 0.05838 | 0.05232 | 0.02440 |

**Table 7.** ANN results

| Peak Training Accuracy | Peak Testing Accuracy |
|---|---|
| 0.9589 | 0.9248 |

## 5. Discussion and Analysis

In comparison to [12], the additional 30 users in this dataset did not seem to have a drastic negative affect on the performance of random forests. In fact, average accuracy increased with this dataset, which can be seen in Table 5 as only 3 of the 10 users from the original dataset were top performers. Moreover, the random forest observed significantly lower FNRs than reported in [12]. Some users' data appear to be more decipherable to one model as opposed to another. For example, user 8 retained top results for the RF and KNN, however is not even in the top-10 for the remaining models. The F1 scores also indicate that that FPR and FNR values at the 0.50 threshold are lower per user on this dataset as compared to the original. The standard deviation of accuracy for random forests was 0.01912, the lowest among machine learning models and lower than what was observed in [12]. Both the 1D-CNN and LSTM-RNN observed miniscule standard deviations among the top-10 users, indicating consistent authentications when model is performing at a high level, but this cannot be extrapolated to all users. Random forests appear to be one of the generally best performing machine learning algorithms for user authentication on mouse dynamics in accordance with the previous literature [34, 35], and our machine learning models' results concur.

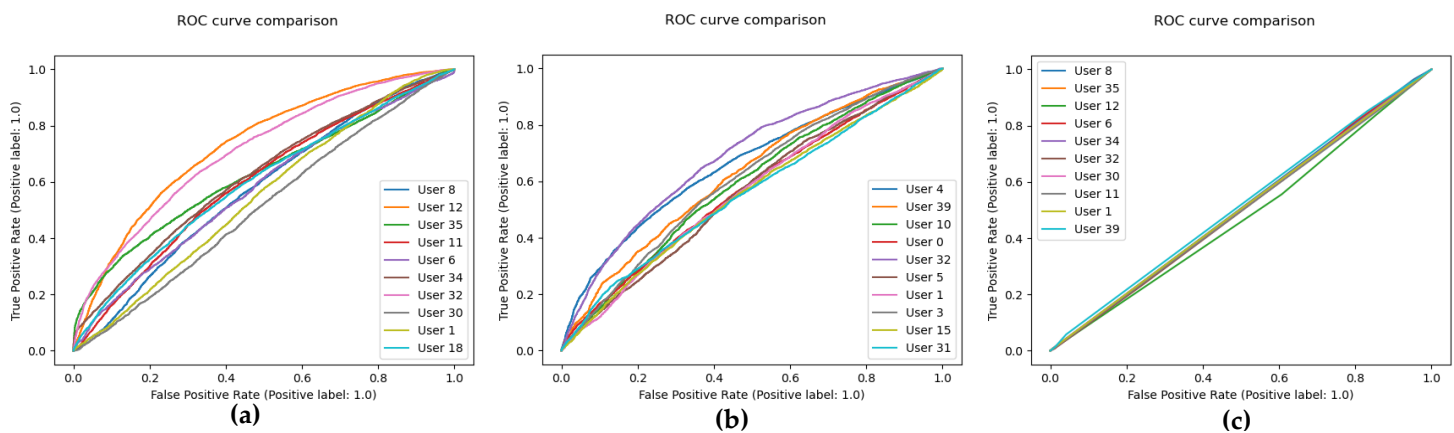

**Figure 3.** (**a**) RF ROC curve; (**b**) SVM ROC curve (**c**) KNN ROC curve

The SVM and KNN had lower performances on the dataset as compared to the random forest. This is most prevalent and easily seen in the ROC curves provided in Figure 3, where the KNN exhibited poor performance. This may prove that the weakness of KNNs with high dimensional data could be too difficult to overcome, but must be further investigated with future research. Compared to SVM results in [30] and in [36], our SVM outperformed both proposed models with lower F1 scores, on our dataset compared to both the Balabit and TWOS dataset. We hypothesized in [12] that collecting data in a faster-paced environment may encode more information in the data as opposed to collecting data from users during dull, administrative tasks. [25] evaluated a deep learning 1D-CNN model on the Balabit dataset while using both higher mouse action block lengths and transfer learning. As opposed to using randomly initialized weights when starting to train a deep learning algorithm, transfer learning allows for the weights of a previously trained model to be initialized instead. This not only speeds up the training process but can also lead to more optimal loss convergence and in turn better performance. Despite these high-level learning concepts utilized for the model in [25], both our SVM and KNN outperformed this model in accuracy, further indicating that our dataset may be of higher quality and better representative of real-world entropy in mouse behavior.

With regards to deep learning, every deep learning algorithm in this paper outperformed our previous random forest in [12]. Deep learning was mentioned in the limitations of [12] and was an important advancement we wanted to address in this experiment. Based on the average accuracy rates of ~85% from our 1D-CNN and LSTM-RNN, we could conclude that deep learning does perform at a higher level on this dataset when compared to simpler machine learning models. Furthermore, our ANN produced an astounding test accuracy of 92.48%; the highest we have seen for this dataset. It is even more remarkable when considering the fact that the ANN is performing in the arguably more difficult multi-class classification setting. [37] did propose an ANN for multi-modal user authentication, using both mouse dynamics and tracked eye-movements, yet still underperformed when compared to our ANN. Moreover, their dataset only contains 32 users. The authors of [37] use various evaluation scenarios to gather a broader perspective of the performance of their model, yet every scenario uses less distinct classes than the 40 used in this article. More importantly, they hand extract features from the raw data before inputting it to their model, whereas our ANN requires no manual feature extraction. Our ANN still outperformed theirs with regards to F1 scores in almost every evaluation scenario. Regardless of F1 scores, our ANN also outperformed the proposed models observed in [38, 39] in raw accuracy rates as well. Note that [39] uses significantly larger mouse action block sizes, ranging from 500 to 2,500. Our dataset is not large enough for such block sizes, but with results already so similar with our block size of 10, it can be concluded that our models are able to operate at similar levels with less available and descriptive data. A summarized version of our results as compared to previous literature can be observed in Table 8.

**Table 8.** Comparison of previous literature and our novel contributions

| Reference | Strengths | Weaknesses | Our Contribution |
|---|---|---|---|
| [12] | - Novel application of RF to our dataset | - Lacks additional models as general baseline references | - Five additional models for application to our dataset<br>- Additional 30 users added to dataset |
| [25] | - One of few papers to explore 1D-CNN application to mouse dynamics | - Use of Balabit dataset limits results to only low-intensity environments<br>- Large mouse action lengths require more data and increase model training times | - Evaluation of 1D-CNN on our high-intensity mouse dynamics dataset<br>- Smaller mouse action lengths decreases data complexity and increases granularity |
| [30] | - Explores various CNN and RNN architectures and their application to mouse dynamics | - Balabit and TWOS dataset use once again limits results to low-intensity environments<br>- SVM observed insufficient results with small mouse action lengths | - Exhibit efficiency of RF and SVM on smaller mouse action lengths due to high-intensity environment<br>- Introduce application of an LSTM-RNN as opposed to GRU-RNN for mouse dynamics |
| [37] | <ul><li>Novel multimodal user authentication using both mouse dynamics and eye movements</li><li>Extensive hyperparameter optimization</li></ul> | <ul><li>Extremely weak ANN results when using mouse dynamics to authenticate 32 users</li><li>Use of eye movements requires more intrusive data collection methods</li></ul> | <ul><li>Our ANN strongly outperformed [37] in all aspects and evaluation metrics</li></ul> |
| [38] | <ul><li>Strong SVM and ANN results</li><li>Uncommon extracted features that lead to promsing results could lead to widerspread adoption</li></ul> | <ul><li>Manual feature extraction required for ANN</li><li>Limits mouse movements to a small interactive GUI for users</li></ul> | <ul><li>Stronger ANN results on dataset is collected in a larger, freeform environement that allows for user behavior that more closely reflects real-world behaviors</li><li>ANN operation does not require manual feature extraction</li></ul> |
| [39] | <ul><li>Novel noise reduction methods based on different aspects of mouse dynamics data</li><li>Novel application of Gaussian Naïve Bayes classifier to dataset proposed in [32]</li></ul> | <ul><li>Vague experimental discussion may lead to irreproducible results</li><li>Few models tested to provide broad baseline references</li></ul> | <ul><li>More detailed description of experimental process aids in fostering future research</li><li>Aforementioned small action block lengths lead to better results with fewer data</li><li>Greater number of models provides better understanding of our dataset's capabilities and shortcomings</li></ul> |

It appears that the added users in the dataset compared to [12] actually increased overall performance and indicates that larger sample sizes in the future could further yield better results. All of our models, to some extent, outperformed others proposed in previous literature, with the most obvious improvements from our previous paper being the introduction of deep learning and the drastically reduced FNRs in the random forests. While this does not guarantee that our models would be able to perform at a higher level if evaluated in a real-world environment, it does indicate that our novel mouse dynamics dataset is sufficient in quality and could yield even better results with future research.

## 6. Limitations

Limitations are important to address, especially in a relatively niche area like mouse dynamics, to encourage continued research and developments. The most glaring improvement to be made is the performance of SVMs and KNNs on the dataset. A moderate range of hyperparameter values were tested for both models, so drastic increases in accuracy from hyperparameter tuning may not be observed for the SVM and KNN. Alternative preprocessing methods or simply more data could lead to better accuracies. For example, many classification problems can also be approached using regression techniques, and mouse dynamics is not an exception [40]. Further exploring the varied user frequency in the top-10 accuracies for each model could also reveal hidden aspects of the data that can be employed to further improve results.

Deep learning architectures were kept quite simple to reduce training time and avoid additional complexities. Increasing the capacity of any of the deep learning models could lead to increased accuracy and F1 scores, possibly at the cost of time. Lastly, the relative novelty of mouse dynamics still leaves plenty of room for future significant breakthroughs. ANNs have been shown in the literature and this article to perform sufficiently well on mouse dynamics data, yet they are quite a general and elementary application of deep learning. When tailored specifically for a certain problem space, stronger and more complex deep learning models than the ones proposed in this article that excel with time-series data could be found to perform at a higher level as opposed to more generally applicable models.

## 7. Conclusions

This article builds upon previous work on a proposed novel dataset for user authentication through mouse dynamics. The previous article tested only one type of machine learning binary classifier on a dataset of 10 users. This article introduces a larger dataset of 40 users and provides three deep learning and three machine learning models as general baselines. Building upon the sole RF proposed in [12], we also evaluate the examine the efficacy of SVMs and KNNs for user authentication using mouse dynamics. The RFs proposed in this article observed generally higher performances than seen in [12] with higher average accuracy, and more evenly distributed FPR and FNR. While the KNNs seemed to struggle on our dataset, our SVMs also observed more evenly distributed FPR and FNR than the RF in [12]. Further, we also introduce a new multi-class classification scenario. The deep learning models expectedly outperformed the machine learning models by an average of around 20% in accuracy, with the ANN performing at an astonishing 92+% test accuracy. This is especially promising as the need for binary preprocessing and classification is unnecessary with the ANN multi-class classification approach. The 1D-CNN and LSTM-RNN were included to evaluate and contrast the different possible approaches to time-series data. Similar accuracies and F1 scores were observed, with the 1D-CNN retaining a slight edge on F1 scores over the LSTM-RNN. A plethora of new avenues could be explored with this dataset and with mouse dynamics as a whole; sustained research is paramount as very promising results have been observed in this article and could continue to appear.

For example, mouse dynamics has really only been developed with low-intensity tasks in mind, such as online banking or administrative tasks, as reflected in the sparsely available public datasets. The introduction of high-intensity mouse dynamics evaluation opens the doors for even more robust user authentication scenarios. With the meteoric rise of professional gaming, user monitoring to prevent cheating or any unfair advantage has become of the utmost importance. Games that require multiple rapid, accurate mouse movements, as seen in many first-person shooters, have observed the unfair use of computer-assisted aiming programs. Models developed on low-intensity mouse dynamics tasks may not be sufficient enough in identifying and flagging such erratic behavior, hence the need for broader mouse dynamics research. Additionally, as covered in the Limitations section, deeper models, alternative preprocessing methods, transfer learning [41], and/or alternative model architectures could all be explored in order to yield better results.

Continuous user authentication using mouse dynamics must be a robust, flexible, and accurate system for widespread adoption to take place. The non-intrusive nature of mouse dynamics paired with the high-performance rate of our models on large, diverse data exhibits the immense potential of this behavioral biometric. Both authentication and classification scenarios in this article are proven to be viable with this dataset and with mouse dynamics as a whole, fostering further research.

**Supplementary Materials:** Any additional information regarding model architecture, implementation or data preprocessing methods can be provided by contacting the corresponding author.

**Data Availability Statement:** The raw data dataset and manually extracted features dataset for all 40 users can be found here: https://github.com/NyleSiddiqui/Minecraft-Mouse-Dynamics-Dataset

**Acknowledgments:** Special thanks to the High-Performance Computing Cluster at UWEC, specifically for the generous access to the BOSE supercomputing cluster for training the models in this article.

**Conflicts of Interest:** The authors declare no conflict of interest.